\documentclass[11pt,a4paper]{article}
\usepackage{times,latexsym}
\usepackage{url}
\usepackage[T1]{fontenc}

\usepackage[acceptedWithA]{tacl2021v1}

\usepackage{xspace,mfirstuc,tabulary}
\usepackage{times}
\usepackage{latexsym}
\usepackage{booktabs}
\usepackage{float}
\usepackage{enumitem}
\usepackage{amsmath}
\usepackage{amssymb}
\usepackage{hyperref}
\usepackage{subcaption}
\usepackage{booktabs}
\usepackage{placeins}

\usepackage{microtype}
\usepackage{footmisc}

\usepackage{inconsolata}
\usepackage{graphicx}

\makeatletter
\AddToShipoutPicture{
  \AtPageLowerCenter{
    \raisebox{1.15cm}{%
      \makebox[0pt][c]{%
        \begin{tabular}{c}
          \footnotesize Accepted for publication in Transactions of the Association for Computational Linguistics (TACL).\\[-0.15em]
          \footnotesize Pre-MIT Press publication version.
        \end{tabular}%
      }%
    }%
  }%
}
\makeatother

\DeclareMathOperator*{\argmax}{argmax}

\title{Emergent Communication in Continuous Worlds:\\ Self-Organisation of Conceptually Grounded Vocabularies at Scale}

\author{
  J\'{e}r\^{o}me Botoko Ekila$^\dagger$ \hspace{0.5cm}
  Lara Verheyen$^\dagger$ \hspace{0.5cm}
  Jens Nevens$^\dagger$ \\
  \textbf{Katrien Beuls}$^\diamond$\footnotemark[1] \hspace{0.5cm}
  \textbf{Paul Van Eecke}$^\dagger$\footnotemark[1]
  \\ \ \\
  $^\dagger$
  Artificial Intelligence Laboratory, Vrije Universiteit Brussel, Belgium \\
  \texttt{\{jerome, lara.verheyen, jens, paul\}@ai.vub.ac.be} \\
  $^\diamond$
  Faculté d’informatique, Université de Namur, Belgium \\
  \texttt{katrien.beuls@unamur.be}
}

\date{}

\begin{document}
\maketitle

\begingroup
\renewcommand{\thefootnote}{\fnsymbol{footnote}}
\footnotetext[1]{Joint last authors. \\This paper was conceived and written without the use of generative AI tools.}
\endgroup

\begin{abstract}
This paper introduces a general methodology through which a population of autonomous agents can converge on a linguistic convention that enables them to refer to arbitrary entities in their environment. The linguistic convention emerges in a decentralised manner through local communicative interactions between pairs of agents drawn from the population. The emergent convention consists of associations between symbolic labels (word forms) and subsymbolic concept representations (word meanings) that are grounded in a continuous feature space. We confirm the generality and scalability of the method through its evaluation on a wide and diverse selection of 37 publicly available datasets. Through a range of experiments, we demonstrate the robustness of the method against perceptual variation, including in heteromorphic populations, as well as the ability of the emergent conventions to self-adapt to changes in the environment.
\end{abstract}

% ----------------
% + INTRODUCTION +
% ----------------
\section{Introduction}

Human languages are evolutionary systems, which emerge and evolve through local communicative interactions between members of a linguistic community. Processes of variation and selection are at play during each and every communicative interaction, at the level of concepts, words and grammatical structures \citep{schleicher1863darwinism, darwin1871descent,maynardsmith1999origins,oudeyer2007language,steels2018evolutionary}. Variants are introduced as creative solutions to communicative impasses and are selected for based on their linguistic, cognitive and physical fitness \citep{grice1967logic,echterhoff2013role}. 

The evolutionary and self-organising nature of human languages gives rise to a number of unique qualities. First of all, such decentralised, self-organising systems are known to be robust and able to self-repair substantial perturbations \citep{heylighen2001science,pfeifer2007self}. Second, populations of language users converge on shared conventions that remain adaptive to changes in their environment and communicative needs \citep{five2009language}. Finally, the resulting languages serve as an abstraction layer above the sensory observations and internal mental representations of individual language users \citep{nevens2020continuous,beuls2024humans,garside2025origin}. Indeed, while linguistic forms can be observed and shared, their meanings remain tied to each language user's individual physical and cognitive embodiment.

This agent-based and evolutionary perspective on the human ability to communicate through language has served as a starting point for the development of a range of computational methodologies that model how artificial agents can co-construct emergent languages that satisfy their communicative needs \citep[see e.g.][]{steels2005coordinating,beuls2013agent,foerster2016learning,lazaridou2017multi,mordatch2018emergence,chaabouni2021communicating, chaabouni2022emergent,nevens2022language,doumen2023modelling, lian2024nellcomx}. Rather than modelling the learning of an existing natural language, these methodologies allow for \textit{artificial natural languages} to emerge and evolve to optimally support the embodiment, environment and communicative needs of populations of artificial agents. These languages are artificial in the sense that they do not exist outside the experimental set-up, yet natural in the sense that they emerge and evolve through the same evolutionary principles as human languages do.

In this paper,  we focus on the emergence of linguistic conventions that associate symbolic labels (referred to as \textit{word forms}) to subsymbolic concept representations (referred to as \textit{word meanings}). We introduce a methodology through which a population of autonomous agents tasked with verbally referring to entities in their environment can converge on a conceptually grounded vocabulary that is adequate for solving their reference task.  The linguistic convention emerges in a decentralised manner through local, task-oriented and situated communicative interactions that take place between pairs of agents drawn from the population.  Importantly,  the entities in the environment of the agents do not come pre-categorised,  but are perceived by the agents as points in a multi-dimensional feature space. As they take part in situated communicative interactions, the agents gradually converge on a vocabulary that associates shared word forms with internal concept representations that are personal yet compatible on a communicative level.

The method and experiments presented in this paper primarily distinguish themselves from earlier work through their focus on how local communicative interactions in fully decentralised populations can lead to emergent global conventions that are not only effective at solving their communicative task, but, crucially, are also robust against perceptual variation, morphological differences between agents, and changing environmental conditions. The evaluation of the method on a diverse selection of 37 publicly available datasets, ranging from physico-chemical analyses to real-world images, confirms at scale the generality of the method across environments. Both the learning process and the emergent languages are to a large extent transparent and human-interpretable, allowing the human experimenter to trace the gradual formation and alignment of conceptual systems, including the formation of niches that structure the conceptual space.\footnotemark

\footnotetext{The source code for the experiments is publicly available at \url{https://gitlab.ai.vub.ac.be/ehai/self-org-at-scale-tacl}.}

\section{Background and Related Work}
\label{sec:related-work}

\paragraph{Usage-based linguistics and constructionist approaches to language.}
From a usage-based perspective, linguistic knowledge is viewed as emerging from situated communicative interactions and grounded in the learner's experience \citep{tomasello2003constructing,bybee2010language,beuls2024humans}. In constructionist theories, such knowledge is formalised in terms of constructions, i.e. pairings of form and meaning, shaped by usage \citep{fillmore1988regularity,goldberg1995constructions,goldberg2006constructions,croft2001radical,langacker2000dynamic,traugott2013constructionalization}. Our experiments reflect this perspective in that agents start without a predefined lexicon or ontology and progressively develop grounded form-meaning pairings through interaction.

\paragraph{Models of language evolution.}
Computational models of language evolution have been used to study how communication systems can emerge and evolve. Early work has analysed signalling systems as equilibria in coordination games  \citep{lewis1969convention,skyrms2010signals}. Later models studied the emergence of conventions in populations of embodied agents, both in simulation and in robotic systems \citep{batali1998computational,cangelosi1998emergence,steels1996perceptually,kirby2001spontaneous,baronchelli2012consensus}. This line of research treats language as a culturally evolving system shaped by learning and interaction.  Within this line of research, \textit{iterated learning} models examine the \textit{vertical transmission} across generations of language learners \citep{kirby2002learning,smith2003iterated,ren2020compositional}, whereas \textit{language game} models focus on \textit{horizontal transmission}, where linguistic conventions emerge through repeated interactions among agents within a population \citep{steels1996perceptually,deboer2001origins,oudeyer2006self,steels2012self}.

\paragraph{Language game models.}
Our experiments and methodology are rooted in the experimental framework of language games \citep{steels1996perceptually,nevens2019practical}. The basic mechanisms were established through experiments on the emergence of grounded naming conventions  \citep{steels2012grounded,loetzsch2015lexicon,steels2016boy},  later moving to grounded concept learning in categorical environments \citep{wellens2008flexible} and in domain-specific continuous environments \citep{steels2005coordinating,bleys2015language,spranger2016referential}.  Further experiments have also modelled the grounding of predefined concepts in perceptual data  \citep{spranger2016referential,wang2016learning,nevens2020continuous}. A key limitation of prior language game experiments concerns the reliance on strict assumptions about perceptual inputs,  which has constrained the scale and complexity of the experimental set-ups and has largely confined this line of work to relatively simple, perceptual domains, such as colour or spatial categories.

\paragraph{Neural emergent communication.}
Models of emergent communication with agents modelled as neural networks trained with reinforcement learning have addressed some of the limitations in scale and complexity of earlier language evolution experiments \citep{foerster2016learning,havrylov2017emergence,kottur2017natural,mordatch2018emergence,kharitonov2019egg,noukhovitch2021emergent,chaabouni2022emergent}.  These models have studied a variety of tasks, including reference games, in more realistic high-dimensional perceptual domains such as real-world images \citep{lazaridou2017multi}. 

A central question concerns the conditions under which robust and generalisable communication systems emerge. One such condition is the presence of populations rather than isolated agent pairs. Several experiments have examined how population size and interaction structure affect emergent languages \citep{sukhbaatar2016learning,das2019tarmac,raviv2019larger,kim2021emergent,chaabouni2022emergent,michel2023revisiting,rita2022role,lian2024nellcomx,lee2024onetomany}, though the experimental findings remain mixed and dependent on task design and agent assumptions.

Another key factor is whether agents are restricted to fixed communicative roles. Most work assumes a division between speakers and listeners, which does not only depart from the interchangeability of speaker and listener roles in human language \cite{hockett1960origin}, but also from the view of communication as a fundamentally interactive, bidirectional process of joint action \citep{clark1996using}. While some experiments have equipped agents with both production and interpretation capabilities \citep{choi2018compositional,portelance2021emergence,nikolaus2024emergent,wolff2024bidirectional}, these are typically restricted to two-agent set-ups. Two notable exceptions have explored bidirectional communication in populations \citep{cao2018emergent,graesser2019emergent}, though they differ in focus: the former focuses on a competitive setting (negotiation) rather than a cooperative one (reference), while the latter involves reference but agents were exposed to a supervised signal.

Finally, much of the neural literature has investigated a particular set-up of the reference game, in which a speaker describes a target entity and a listener selects it from a set of `distractors' based on the utterance \cite[e.g.][]{havrylov2017emergence,portelance2021emergence,chaabouni2022emergent}. Although this setting has proven effective for studying the emergence of language, it abstracts away from the broader situational context in which communication occurs. Recent work has shown that incorporating contextual information improves communicative efficiency \citep{gualdoni2024bridging,glowka2024contextdependent,zhang2025nellcomlex} as it allows speakers to choose concepts with greater granularity \citep{ohmer2022emergence,kobrock2024context}. However, these experiments have so far only examined two-agent set-ups.

% ----------------------
% + PROBLEM DEFINITION +
% ----------------------
\section{Problem Definition}
\label{sec:problem}

We address a decentralised, multi-agent emergent communication problem in which agents must bootstrap a linguistic convention that allows them to draw each other's attention to arbitrary entities in their environment.  Importantly,  communicative interactions take place locally between pairs of agents, agents can both act as speakers and  listeners,   the environment does not come pre-categorised,  and the emergent convention needs to be suitable for communication about previously unseen entities. The problem definition thereby brings together a number circumstances under which human language emerge and evolve, which have often been studied in isolation in prior work.  More formally, the problem can be defined as follows:

\paragraph{Population.}
There exists a population $P = \{a_1, \ldots, a_n\}$ that consists of $N$ autonomous agents.   Agents have no access to each other's internal state nor to any centralised knowledge base,  and start out as `blank slates' without any words,  concepts or knowledge about the world.  

\paragraph{World.}
There exists a world $W = \{e_1, \ldots, e_K\}$ that consists in a set of $K$ entities.  An observation of an entity by an agent $a$ takes the form of a feature vector $X_a$ of $d$ dimensions,  for example resulting from the agent's sensor read-outs.  The dimensions of such a vector can be continuously-valued, categorically-valued or a combination of both.  Importantly, every agent perceives the environment through its own sensors, so it cannot be assumed that all agents perceive a given entity identically or even represent it as a vector of the same dimensionality.

\paragraph{Interactions.}
Agents take part in a sequence of task-oriented communicative interactions.  At the beginning of each interaction,  a scene $C = \{e_1, \ldots, e_k\} \subset W$ of $k$ entities from the world is randomly created.  Two agents are randomly selected from $P$. One is assigned the role of speaker ($S$) and the other the role of listener ($L$).  A topic entity $T \in C$ is randomly selected from the scene and is only disclosed to $S$.  $S$ is tasked with drawing the attention of $L$ to $T$ by producing an utterance $U$ that is passed on to $L$.  $L$ should then identify $T$.  Success occurs if $L$ correctly identifies $T$.  In case of failure,  $T$ is disclosed to $L$.  After the interaction,  both agents are informed about whether the interaction succeeded or failed.  Identification or disclosure of entities always happens in terms of the agents' own perceived feature vectors,  i.e.  $X_S$ for $S$ and $X_L$ for $L$. 

The formal definition of the problem is deliberately generic and can be straightforwardly instantiated in a variety of scenarios. For example, in a robotic scenario, agents may be equipped with sensors, in which case the perceived feature vector corresponds to sensor readings for a given entity. In simulated settings, scenarios could involve populations of simulated agents communicating about entities that are stored as entries in tabular datasets.  In such cases,   agents `perceive' a given entry  as the vector composed of that entry's (normalised) column values. In visual domains, agents may communicate about high-dimensional perceptual inputs derived from real-world images, where each image is treated as a single entity.

% --------------
% + METHDOLOGY +
% --------------
\section{Methodology}
\label{sec:methodology}

The methodology that we present is grounded in usage-based and constructionist theories of language, in which linguistic structure emerges through situated communicative interactions and is shaped by usage \citep{tomasello2003constructing,beuls2024humans}.  From this perspective, referential communication is a collaborative process: interlocutors coordinate on an intended referent in context \citep{clark1986referring,clark1996using,pickering2004toward}. This coordination relies on the gradual accumulation of knowledge in the form of constructions, that is, form-meaning pairings, with each individual building up their own inventory over time.  When existing constructions do not suffice, speakers may introduce new ones  to resolve communicative impasses, which through repeated use become progressively more conventionalised and entrenched \citep{bybee1998emergent,bybee2010language,langacker2000dynamic}. Our methodology computationally operationalises these general principles with regard to the problem definition, building on prior language game research \cite{steels1995self,wellens2008flexible,nevens2020continuous}.

\paragraph{Linguistic inventory.}
Each agent $a \in P$ has its own linguistic inventory $I_a$, which is initially empty. Every word $w \in I_a$ is a triple $w = (f,c,s)$ consisting of a word form $f \in F$, a concept representation $c$ and an entrenchment score $s \in [0,1]$. $F$ is an unbounded set of possible atomic word forms. In the experiments, forms are generated as atomic three-syllable consonant-vowel strings (e.g. ``\textit{demoxu}''). This generation procedure is arbitrary and chosen for ease of inspection. Word forms on their own thus carry no semantic information and play no role in the learning dynamics.

\paragraph{Concept representations.}
A concept representation $c = ((\omega_1, \theta_1) \ldots (\omega_d, \theta_d))$ consists of a pair of a weight $\omega$ and a distribution $\theta$ for each dimension of the feature vector $X$ perceived by the agent. For continuous dimensions, $\theta_i$ is a normal distribution parametrised by $(\mu_i, \sigma_i)$. For categorical dimensions, $\theta_i=\mathrm{f}_i$ is a count vector over the $k$ possible categories, where $\mathrm{f}_{i,j}$ denotes the absolute frequency of category $j$. The weight $\omega_i$ represents the importance of dimension $i$ for the concept. To evaluate how well a concept applies to a perceived entity, the similarity between a concept~$c$ and $X=(x_1,\ldots,x_d)$ is computed dimension-wise and aggregated with normalised weights:

\begin{equation}
  \small
  \mathrm{sim_{c\textrm{--}e}}(c, X) = \sum_{i=1}^{d} \frac{\omega_i}{\sum_{k=1}^d \omega_k} \mathrm{sim_d}(\theta_i, x_i)
  \label{eq:concept-similarity}
\end{equation}

\begin{equation}
  \small
  \mathrm{sim_d}(\theta_i, x_i) = 
  \begin{cases}
  \exp\left(-\left|\frac{x_i - \mu_i}{\sigma_i}\right|\right)&\text{if continuous dim.} \\
  \frac{\mathrm{f}_{i,x_i}}{\sum_{j=1}^{k} \mathrm{f}_{i,j}} &\text{if categorical dim.} 
  \end{cases}
  \label{eq:dimension-similarity}
\end{equation}

Normalising the weights avoids a bias towards concepts that distribute relevance across more dimensions. 

Concepts therefore define prototypical categories grounded in the agent's own perceptual space \citep{rosch1973internal} and are iteratively refined as agents take part in communicative interactions. These representations serve as the basis for producing and interpreting utterances.

\paragraph{Conceptualisation and production.}
Given a scene $C$, a topic $T \in C$ and speaker $S$, $S$ first retains only the words whose concepts discriminate $T$ from all other entities in $C$:

\begin{equation}
  \small
  K = \{ w_i \in I_S \mid \mathrm{sim_{c\textrm{--}e}}(c_i, T) > \max_{e\in C \setminus T}\mathrm{sim_{c\textrm{--}e}}(c_i, e)\}
  \label{eq:candidate-words}
\end{equation}

If $K \neq \varnothing$, the speaker $S$ selects the candidate $w^*$ with the highest \textit{communicative adequacy}, defined as  the product of its current score $s$ and its \textit{discriminative power}, i.e. its ability to discriminate $T$ from the other entities in $C$:

\begin{equation}
  \small
  w^* = \argmax_{w_i \in K} s_i \cdot \left[\mathrm{sim_{c\textrm{--}e}}(c_i, T) - \max_{e\in C \setminus T}\mathrm{sim_{c\textrm{--}e}}(c_i, e) \right]
  \label{eq:select-best-candidate}
\end{equation}

The word form of $w^*$ is uttered by $S$ as $U$ to the listener $L$.

\paragraph{Invention.}
If $K = \varnothing$, the speaker $S$ faces a communicative impasse and invents a new word $w = (f,c,s)$.  The new form $f$ is sampled from $F$, the concept $c$ is initialised from  the speaker's percept $X_S$ of the topic, and the word is added to $I_S$. The form $f$ is uttered as $U$.

Whenever an agent $a$ creates a new word $w = (f,c,s)$ in its inventory $I_a$ from a perceptual vector $X_a$, the concept $c$ is initialised as follows.  For continuous dimensions, $\mu_i$ is set to $X_i$ and $\sigma_i$ to the default value $\sigma_{ini}$.   For categorical dimensions, the observed category in $X_i$ is assigned a frequency of $1$.  All weights and the word score are initialised to the fixed values $\omega_{ini}$ and $s_{ini}$.

\paragraph{Comprehension and interpretation.}
Upon observing $U$, the listener checks whether it knows a word with that form. If so, let $w = (U,c,s) \in I_L$,  $L$ identifies the entity in the scene $e \in C$ that is most similar to $c$ as the hypothesised topic $T^*$:

\begin{equation}
  \small
  T^{*} = \argmax_{e_i \in C} \mathrm{sim_{c\textrm{--}e}}(c, e_i)
  \label{eq:interpretation}
\end{equation}

If $T^*=T$, the interaction is considered successful.  Otherwise it fails, and $T$ is disclosed to the listener $L$.  After each communicative interaction,  both $S$ and $L$ will update the words and concept representations in their respective linguistic inventories $I_S$ and $I_L$.  We distinguish between successful interactions and failed interactions.

\paragraph{Successful interaction update.}
Successful use reinforces the selected word and inhibits competing alternatives. Let $w_U=(U,c_U,s)$ be the selected word.   This word receives a fixed reward $s_r$, while competing candidates, i.e. the words in the candidate set $K$ excluding $w_U$, are penalised in proportion to their similarity to $c_U$:

\begin{equation}
  \small
  s \leftarrow s +
  \begin{cases}
    s_r & \text{if } w = w_U \\
    s_{li} * \mathrm{sim_{c\textrm{--}c}}(c, c_U) & \text{if } w \in K \setminus w_U
  \end{cases}
  \label{eq:score-update}
\end{equation}

Thus, words that are more similar are considered stronger competitors and are punished harder. The similarity between two concept representations ($\mathrm{sim_{c\textrm{--}c}}$), where $D_f$ is the f-divergence between the two parametrised distributions \citep{hellinger1909neue}, is defined as follows:

\begin{equation}
  \small
  \begin{split}
  \mathrm{sim_{c\textrm{--}c}}(c_q, c_r)& = \sum_{i=1}^{d}\underbrace{\left( 1 - D_{f}{(\theta_{q,i} \parallel \theta_{r,i})}\right) }_{\text{distribution similarity}}\\
  & * \underbrace{\left(1 - |\frac{\omega_{q,i}}{\sum_{k=1}^{d}\omega_{q,k}} - \frac{\omega_{r,i}}{\sum_{k=1}^{d}\omega_{r,k}}|\right)}_{\text{normalised weight similarity}} \\
  & * \underbrace{\frac{\frac{\omega_{q,i}}{\sum_{k=1}^{d}\omega_{q,k}} + \frac{\omega_{r,i}}{\sum_{k=1}^{d}\omega_{r,k}}}{2}}_{\text{average normalised weights}}
  \end{split}
  \label{eq:concept-concept-similarity}
\end{equation}

The listener $L$ applies the same score update in its own inventory $I_L$, by collecting all words in $I_L$ that it would consider candidates and constructs its own set $K$ based on its own perceptual view of the scene.

After a successful interaction, both agents also update the concept $c$ associated with $U$. For continuous dimensions, $\mu_i$ and $\sigma_i$ are updated online using Welford's algorithm \citep{welford1962note}. For categorical dimensions,  the observed category count is incremented. Dimension weights are updated according to discriminative power in the current scene:  dimensions with  positive discriminative power are rewarded by a fixed step $\omega_r$ on a sigmoid function, while the remaining dimensions are decreased by a fixed step $\omega_p$ on the same function. This bounds weights between 0 and 1, with weights becoming more stable as they approach 0 or 1.

\paragraph{Failed interaction update.}
After a failed interaction,  $S$ will decrease the score of $w_U = (U,c,s) \in I_S$ by a fixed value $s_p$.  If $L$ knew a word with the observed form $U$,  $L$ will decrease  the score of $w_U = (U,c,s) \in I_L$ by a fixed value $s_p$ and update its $c$ based on $T$ relative to $C$ in the same way as if the interaction had been successful.  If $L$ did not know a word $w = (U,c,s)$,  $L$ will adopt the word as follows:

\paragraph{Adoption.}
A new word $w = (f,c,s)$ is added to $I_L$,  with $f$ being the observed utterance $U$ and $c$ initialised based on the listener's percept $X_L$ of the disclosed topic using the word initialisation procedure (see Invention).

\paragraph{Multi-word utterances.}
The methodology can be generalised to multi-word utterances. Instead of requiring a single word to uniquely discriminate the topic $T$ in scene $C$, the speaker maintains an entity set $E$, initially equal to $C$, and iteratively selects words that reduce $E$ while preserving $T$. At each step, candidate words are evaluated with respect to the current set $E$. For a candidate word $w_i$, selecting it updates $E \leftarrow \{ e \in E \mid \mathrm{sim_{c\textrm{--}e}}(c_i,e) \ge \mathrm{sim_{c\textrm{--}e}}(c_i,T) \}$.  Candidate words are ranked first by the reduction they achieve on $E$ and, in case of ties,  by communicative adequacy as defined in Equation~\ref{eq:select-best-candidate}.  The procedure terminates when $E$ contains only $T$ or when the maximum utterance length $\ell$ is reached. The resulting utterance is the sequence of selected word forms. When $\ell=1$, this reduces to the single-word utterance case. When $\ell>1$, the speaker can combine several partially discriminative words into a compositional utterance.  On the listener side, comprehension mirrors production by incrementally reducing the candidate set as words are processed sequentially. Alignment updates are applied stepwise: for each selected word, competitors are the words that achieve the same reduction of the current set $E$, and score updates are computed exactly as in the single-word case with respect to that step-specific competitor set.

% ----------------------
% + Experimental Setup +
% ----------------------
\section{Experimental setup}
We instantiate the problem definition in 37 scenarios and highlight seven prototypical scenarios differing in feature type: continuous (\textsc{clevr}, \textsc{wine}, \textsc{mscoco}, \textsc{celeba}, \textsc{birds}), categorical (\textsc{mushrooms}) and a combination of both (\textsc{exoplanets}). The \textsc{clevr} scenario uses visual features extracted from images in the CLEVR dataset \citep{johnson2017clevr}. \textsc{wine} relies on physico-chemical measurements of wine samples \citep{cortez2009modeling}. \textsc{exoplanets} combines numerical and categorical features of exoplanets \citep{mishra2023nasa}. \textsc{mushrooms} consists of categorical features describing mushroom attributes \citep{schlimmer1981mushroom}. The three real-world images scenarios are \textsc{mscoco} \cite{lin2014microsoft}, \textsc{celeba} \cite{liu2015deep} and \textsc{birds} \cite{wah2011caltech}. Following common practice in the literature \citep{lazaridou2017multi,chaabouni2022emergent}, each image is processed by a pre-trained vision encoder. In line with \citet{kouwenhoven2024curious}, we use DINOv2 \cite{oquab2023dinov2}, yielding a dense vector representation that serves as the agent's perceptual input. Full processing details and dataset specifications for all 37 scenarios are provided in Appendix~B.

In each scenario, the world $W$ is defined as the set of entries from the underlying dataset. We hold out 25\% of the entities in $W$ for testing purposes. At the beginning of each interaction,  a new scene is created by randomly selecting 10 entities from $W$,  with the constraint that training scenes can only hold training entities and that test scenes can only hold (unseen) test entities.  The exception to this rule is  \textsc{clevr},  where the original dataset splits are used, holding scenes of 3 to 10 entities.

In each experiment, unless specified otherwise, we train a population of 10 agents for 1M pairwise interactions on the training scenes and evaluate on 100K interactions on the testing scenes. Results are averaged over 10 independent runs. The chosen hyperparameters are shown in Table~\ref{tab:parameters}. These values have only been tuned on the \textsc{clevr} dataset and are used on all 37 datasets with no further fine-tuning.

\begin{table}
  \centering
  \small
  \begin{tabular}{ccl}
  \toprule
  \textbf{Param.}  & \textbf{Value} & \textbf{Description} \\ 
  \midrule 
  $n$ & $10$ & \# agents in population \\ 
  $k$ & $10$ & \# entities in scene\\ 
  \hline
  $s_{ini}$ & $0.5$ & initial word score\\ 
  $s_r$ & $+0.1$ & word score reward \\ 
  $s_p$ & $-0.1$ & word score punishment \\ 
  $s_{li}$ & $-0.02$ & competitor score punishment\\
  \hline 
  $\sigma_{ini}$ & $0.01$ & initial standard deviation\\ 
  $\omega_{ini}$ & $0.5$ & initial dimension weight\\ 
  $\omega_r$ & $+1$ & dimension weight reward\\ 
  $\omega_p$ & $-5$ & dimension weight punishment\\ 
  \bottomrule 
  \end{tabular} 
  \caption{Overview of parameter settings.}
  \label{tab:parameters}
\end{table}

\begin{table*}[ht]
  \small
  \centering
  \begin{tabular}{lrrrrrrrr}
  \toprule
  \textbf{Dataset} & \textbf{\# ent.} & \textbf{\# cont.} & \textbf{\# cat.} & \textbf{succ. (\%) }& \textbf{conv. (\%)} & \textbf{inv. size} & \textbf{inv. size (95\%)} \\
  \midrule
  \textsc{clevr}            & 468K & 20 & 0 & 99.76$_{\pm 0.08}$ & 94.41$_{\pm 1.46}$ & 52.63$_{\pm 1.73}$ & 31.14$_{\pm 1.11}$ \\
  \textsc{wine}             & 5K & 12 & 0 & 99.76$_{\pm 0.17}$ & 88.34$_{\pm 1.31}$ & 77.12$_{\pm 1.21}$ & 60.25$_{\pm 1.37}$ \\
  \textsc{exoplanets}       & 5K & 8 & 4 & 99.67$_{\pm 0.10}$ & 92.30$_{\pm 0.86}$ & 79.35$_{\pm 1.18}$ & 54.47$_{\pm 1.67}$ \\
  \textsc{mushrooms}        & 8K & 0 & 23 & 98.10$_{\pm 0.31}$ & 86.55$_{\pm 1.82}$ & 265.40$_{\pm 6.77}$ & 76.32$_{\pm 2.05}$ \\
  \midrule
  \textsc{mscoco}           & 159K  & 384   & 0   & 95.77$_{\pm 1.02}$ & 95.81$_{\pm 0.74}$ & 17.63$_{\pm 0.15}$ & 14.67$_{\pm 0.58}$ \\
  \textsc{celeba}           & 203K  & 384   & 0   & 96.66$_{\pm 1.88}$ & 95.60$_{\pm 1.58}$ & 19.93$_{\pm 0.92}$ & 16.20$_{\pm 1.04}$ \\
  \textsc{birds}            & 12K   & 384   & 0   & 95.88$_{\pm 3.11}$ & 96.11$_{\pm 1.29}$ & 19.50$_{\pm 2.71}$ & 15.00$_{\pm 2.00}$ \\
  \bottomrule
  \end{tabular}
  \caption{Experimental results on the test sets of the seven featured scenarios after 1M training interactions. Mean and $\pm$2 standard deviations computed over 10 independent runs.  Columns describe the dataset, number of entities, number of continuous and categorical dimensions, communicative success, conventionality and linguistic inventory size. Results for the remaining 30 datasets are included in Appendix~C.}
  \label{tab:tabular-results}
\end{table*}

In line with common practice in the field \citep{steels1999talking,loetzsch2015lexicon},  the results are analysed in terms of three quantitative metrics both during training and at test time: 

\paragraph{Degree of communicative success.}
The degree of communicative success reflects how successful a population of agents is at solving the task. It is computed as the average outcome of all interactions, where success counts as 1 and failure as 0. It serves as the feedback signal through which agents reinforce or adjust their linguistic structures.

\paragraph{Degree of conventionality.} 
The degree of conventionality quantifies to what extent the different agents in the population would produce the same utterance under the same circumstances,  thereby measuring convergence towards a predictable linguistic convention.  It is computed by averaging, over all interactions, a binary measure that indicates whether the listener agent would have used the same utterance as the one produced by the speaker agent to describe the topic entity, if the listener had been the speaker. In prior work, this metric has been referred to as \textit{lexical coherence} \cite{steels2015talking}. Other formulations for this metric are \textit{message agreement} \citep{kim2021emergent} and \textit{speaker synchronisation} \citep{rita2022role}.

\paragraph{Linguistic inventory size.}
The average linguistic inventory size (\textit{inv.~size}) is calculated as the average number of distinct words uttered by the agents. To estimate the most frequently used words, we quantify the number of distinct word forms required to account for 95\% of all produced utterances (\textit{inv.~size (95\%)}).

% -----------
% + RESULTS +
% -----------
\section{Results}
\label{sec:results}

The experiments address two central questions: whether the proposed methodology scales across diverse perceptual domains, and whether the resulting conventions remain effective when perception differs across agents. We first establish performance in homogeneous populations, then analyse the dynamics of the emergent conventions, before turning to experiments involving perceptual heterogeneity and multi-word utterances.

\subsection{Reference in homogeneous populations}
\label{sec:homogeneous}

We first evaluate the methodology in a base setting with homogeneous populations, assessing its performance across the 37 datasets. Consistent with prior work, we assess zero-shot generalisation to unseen (in-distribution) samples by evaluating the resulting convention on entities not seen during training \cite{kottur2017natural,lazaridou2018emergence}. Table~\ref{tab:tabular-results} reports test set performance on the seven featured scenarios in terms of communicative success, conventionality and linguistic inventory size (including the metric to track the most frequently used words).  In each scenario, the population reaches a degree of communicative success of over 98\%,  with a degree of conventionality above 86\%.  The average linguistic inventory size ranges from 17 to 265 words. A core of 14 to 76 words covers 95\% of usage, indicating a heavy-tailed distribution. The evaluation results on the remaining 30 datasets are included as supplementary materials to this paper, and are in line with those obtained in the seven featured scenarios. These results confirm that the populations consistently converge on communicatively effective and conventional languages with a limited number of words as compared to the number of entities in the training data.

\paragraph{Generalisability.}
Next, we assess the generality of the emergent concepts in terms of their adequacy to refer to entities that exhibit previously unseen attribute combinations, a challenge referred to as \textit{generalisability} \citep{boldt2024review, lazaridou2018emergence, lee2024onetomany}. This experiment can be interpreted as a controlled test of whether the convention remains robust under a shift in the environment, namely in the distribution of attribute combinations. We apply the methodology to the CLEVR CoGenT dataset \citep{johnson2017clevr}, which was especially designed to test the robustness of intelligent systems against correlations that occur at test time but not during training. As such, a number of biases are built into the scenes. For example, in the training scenes, cubes are always grey, blue, brown or yellow. Test set A contains scenes that are subject to the same correlations, whereas test set B consists of scenes that are subject to different correlations. Test set B can be used to assess whether the learnt model generalises beyond the correlations that characterise the training set. The results show that the performance of the agents on test set B closely matches that on test set A, with a communicative success of 99.75\% and 99.78\% respectively. The generalisability experiment thereby confirms that the emerged linguistic convention does not break down when faced with the need to refer to entities that instantiate previously unseen attribute combinations.

\begin{figure}
  \includegraphics[width=\columnwidth]{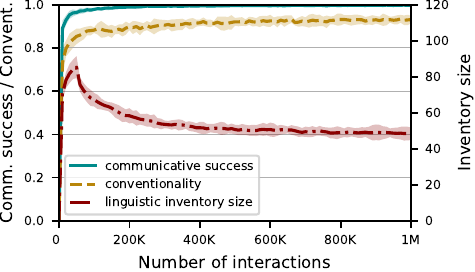}
  \caption{Evolutionary dynamics during the training phase of the \textsc{clevr} experiment with 10 agents. Mean shown as a line, shaded region indicates $\pm 2$ standard deviations over 10 runs.}
  \label{fig:clevr-baseline}
\end{figure}

\paragraph{Evolutionary dynamics.}
The evolutionary dynamics that take place during the training phase of the \textsc{clevr} experiments are visualised in Figure~\ref{fig:clevr-baseline}.  The graph shows the degree of communicative success (solid line, left y-axis),  the degree of conventionality (dashed line, left y-axis) and the average linguistic inventory size (dashdotted line, right y-axis) as a function of the number of communicative interactions that took place and averaged over a sliding window of 5K interactions.  Communicative success rises to about 96\% after 50K interactions, and continues to grow to 99.81\% over the course of the 1M interactions that take place. The degree of  conventionality roughly follows the same dynamics as the degree of communicative success, although the growth is much slower.  After 1M interactions, the degree of  conventionality has reached 93.30\%.  The average linguistic inventory size shows the typical `overshoot pattern' that is found in many language emergence experiments \citep{vaneecke2022language}.  Many words emerge during the initial phase of the experiment, as the individual agents are constantly faced with the need to invent.  Then, as a result of the rewarding and punishing of words, the population converges on a smaller inventory size. The graph shows that the peak linguistic inventory size lies around 90 words, while an average of 48.42 words is reached after 1M interactions.

\begin{figure}[t]
  \includegraphics[width=\columnwidth]{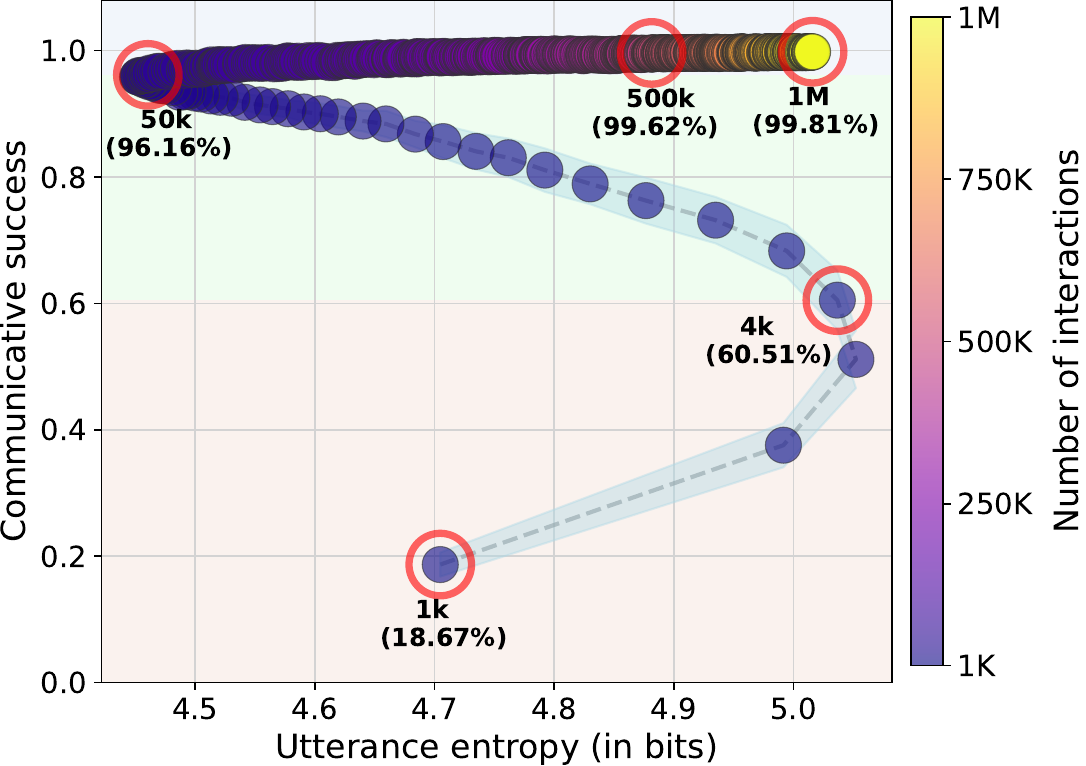}
  \caption{Communicative success plotted against utterance entropy $H(U)$ over training. Every circle aggregates 1000 interactions and colour encodes training progress.
  }
  \label{fig:ib-experiments-clevr}
\end{figure}

\paragraph{Evolution of an emergent inventory.}
We examine how the word usage frequency distribution evolves over time. Figure~\ref{fig:ib-experiments-clevr} visualises the evolution of communicative success against utterance entropy in the \textsc{clevr} scenario. Utterance entropy ${H}(U)$ is computed as the Shannon entropy of the empirical distribution over the agents' utterances. We identify three phases. In an early invention phase, entropy grows fast as agents invent and test out novel words. During an alignment phase, success rises sharply and entropy drops as agents converge on broadly effective but still coarse concept representations. Finally, a slow refinement phase follows, in which entropy rises again. In this final phase, few new words emerge, rather existing ones are gradually tuned or repurposed to fill remaining conceptual gaps. This phase progressively slows down as further gains in communicative success become marginal.

\begin{figure}[t]
    \includegraphics[width=\columnwidth]{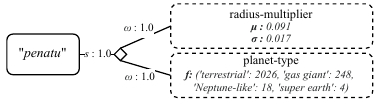}
    \caption{Example of a word emerged in the \textsc{exoplanets} scenario. It has specialised towards two dimensions, one continuous (`radius-multiplier') and the other categorical (`planet-type').}
    \label{fig:words-exo}
\end{figure}

\paragraph{Effect of the population size on convergence.}
We test how the approach scales with varying population sizes. Following \citet{chaabouni2022emergent}, we test populations of up to 100 agents trained for 1 to 10M interactions on the \textsc{clevr} scenario (see detailed results in Appendix~D). We observe similar dynamics as in Figure~\ref{fig:clevr-baseline}, but due to the number of interactions per agent decreasing as population size increases, it takes more interactions for larger population sizes to converge on an effective convention. However, the size of this convention (as measured by the linguistic inventory size) is smaller. We hypothesise that larger populations introduce more variants, allowing selection to favour the most effective variants. Similar findings were reported by \citet{lian2024nellcomx}, who found that smaller populations tend to settle on effective yet sub-optimal, redundant conventions.

\paragraph{Example of an emerged word.}
Figure~\ref{fig:words-exo} shows a word with the form ``\textit{penatu}'' that emerged in agent 1 in the \textsc{exoplanets} scenario and was fully entrenched after 1M interactions ($s=1.0$).  Two dimensions are important in the concept representation of this word ($\omega > 0.0$): the continuously-valued dimension  \textit{radius-multiplier} (expressed in earth radii) and the categorically-valued dimension \textit{planet-type} (e.g. `terrestrial').  De-normalising the \textit{radius-multiplier} value reveals that ``\textit{penatu}'' prototypically refers to terrestrial-type exoplanets with a radius around 81\% of the earth's radius.

\begin{figure*}
  \begin{subfigure}{\columnwidth}
    \includegraphics[width=\columnwidth]{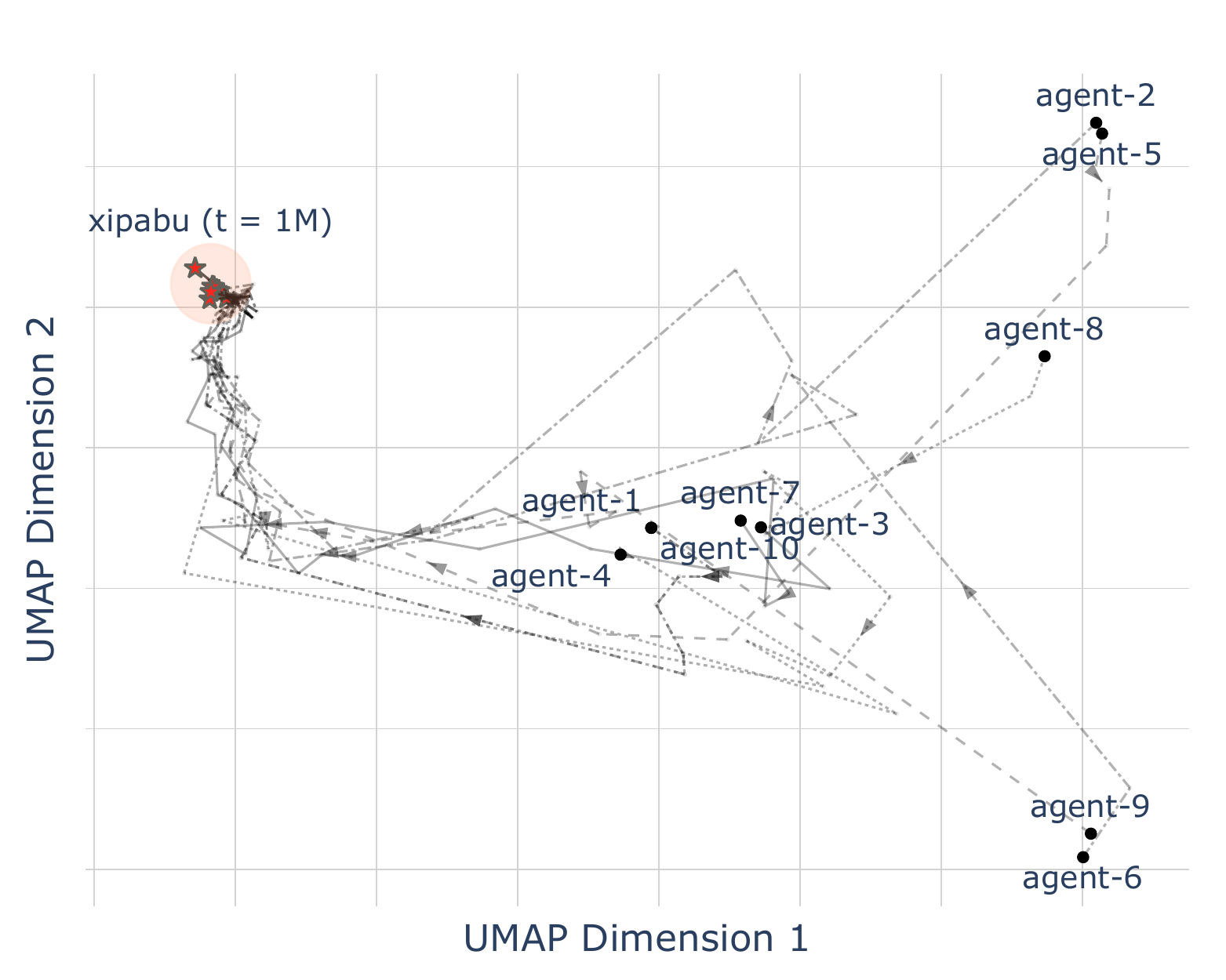}
    \caption{Single word trajectory, 10 agents}
    \label{fig:umap-single-word}
  \end{subfigure}
  \begin{subfigure}{\columnwidth}
    \includegraphics[width=\columnwidth]{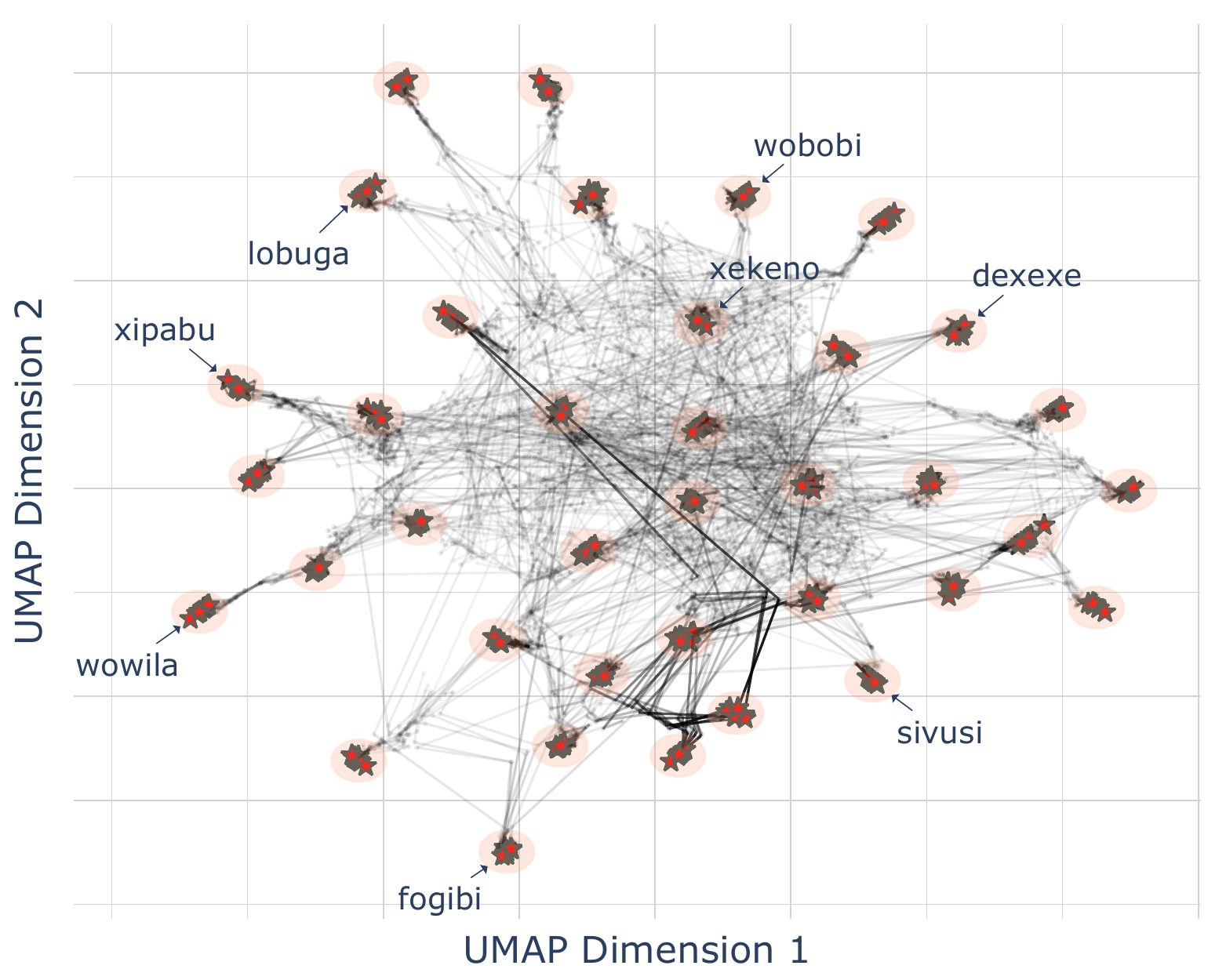}
    \caption{Trajectories of all words with $s>0$, 10 agents}
    \label{fig:umap-lexicon}
  \end{subfigure}
  \caption[UMAP]{Aligned-UMAP visualisations of the trajectories of concept representations over time.}
  \label{fig:umap}
\end{figure*}

\paragraph{Trajectory of an emergent word.}
Figure~\ref{fig:umap} visualises the trajectories that words follow as they are shaped during training. Each word is represented at each time step by the concatenation of its parameters (score, weights and distributional parameters). These representations are projected in two dimensions using the Aligned-UMAP technique for temporal data \cite{mcinnes2018umap}.  Subfigure~\ref{fig:umap-single-word} shows the trajectory of the concept representation associated with the word ``\textit{xipabu}'' in an experiment with 10 agents on the \textsc{clevr} scenario.  Initially,  the concept representations across the 10 agents are very different,  as each was learnt locally from a specific interaction.  Over time,  the representations of the different agents align as a result of the evolutionary dynamics that take place.  Subfigure~\ref{fig:umap-lexicon} shows the trajectories of all words in the final linguistic inventories of the 10 agents. Figure~\ref{fig:umap} not only shows the alignment of concept representations but also the formation of niches that structure the conceptual space. These niches arise as competing words progressively differentiate in meaning as they are used, very much in the spirit of \citet{breal1897essai}'s laws of \textit{spécialité} and \textit{répartition}.

% --------------------------
% + Robustness EXPERIMENTS +
% --------------------------

\subsection{Robustness to perceptual heterogeneity}
\label{sec:robustness}

So far, the experiments have considered homogeneous populations of agents where agents perceive entities through identical input feature spaces. These settings however abstract away from an important property of grounded communication, namely that meanings are tied to each agent's own embodiment. We therefore now turn to conditions under which agents do not perceive entities identically, testing whether successful conventions can still emerge when perception varies across agents or changes over time.

\paragraph{Uncalibrated sensors and noisy environments.}
The first two experiments assess the robustness of the methodology against differences in the agents' perception of the world (i.e. $X_S \neq X_L$). As shown in \citet{chaabouni2022emergent}, even mild input noise, typically simulated by injecting Gaussian noise into agents' inputs, strongly reduces communicative success. In our setup, we simulate two distinct sources of perceptual variation. In a first scenario, a lack of calibration (Ca) is simulated by shifting $X_a$ by a value that is individually sampled at the beginning of each experiment for each sensor of each agent from a normal distribution with a mean of $0$ and a standard deviation of $0.001$ or $0.01$ (Ca1 and Ca2 conditions). In a second scenario, noise (No) is added to the sensor values by shifting $X_a$ by a value that is independently sampled at the beginning of each interaction for each sensor of each agent from a normal distribution with a mean of $0$ and a standard deviation of $0.001$ or $0.01$ (No1 and No2 conditions). 

As seen in Table~\ref{tab:robustness},  a lack of calibration minimally impacts the emergent conventions. Adding random sensor noise does lead to less conventional languages, but the communication does not break down even under substantial amounts of noise.

\begin{table}[h]
  \small
  \centering
  \begin{tabular}{lllll}
  \toprule
  \textbf{Test} & \textbf{succ.} (\%) & \textbf{conv.} (\%) & \textbf{inv. size} \\
  \midrule
  \midrule
  \multicolumn{5}{l}{\textbf{Baseline}} \\
  \midrule
  -     & 99.76$_{\pm 0.08}$ & 94.41$_{\pm 1.46}$ & 52.63$_{\pm 1.73}$ \\
  \midrule
  \multicolumn{5}{l}{\textbf{Uncalibrated sensors and noisy environments}} \\
  \midrule
  Ca1  & 99.77$_{\pm 0.11}$ & 93.05$_{\pm 2.13}$ & 54.45$_{\pm 2.33}$ \\
  Ca2  & 99.74$_{\pm 0.10}$ & 92.41$_{\pm 1.28}$ & 54.95$_{\pm 1.50}$ \\
  No1  & 98.81$_{\pm 0.52}$ & 81.08$_{\pm 3.69}$ & 54.69$_{\pm 1.21}$ \\
  No2  & 87.67$_{\pm 0.91}$ & 42.55$_{\pm 2.67}$ & 57.29$_{\pm 2.20}$ \\
  \midrule
  \multicolumn{5}{l}{\textbf{Heteromorphic populations}} \\
  \midrule
  Ho1 & 99.80$_{\pm 0.07}$  & 93.41$_{\pm 1.08}$  & 52.59$_{\pm 2.09}$ \\
  He1 & 98.12$_{\pm 1.60}$  & 87.42$_{\pm 2.40}$  & 53.73$_{\pm 2.24}$ \\
  Ho2 & 99.72$_{\pm 0.16}$  & 92.39$_{\pm 2.72}$  & 53.42$_{\pm 3.06}$ \\
  He2 & 83.47$_{\pm 11.07}$ & 57.03$_{\pm 17.61}$ & 65.49$_{\pm 3.02}$ \\
  \midrule
  \multicolumn{5}{l}{\textbf{Robustness against sensor defects}} \\
  \midrule
  De1  & 99.22$_{\pm 1.05}$  & 90.57$_{\pm 2.60}$  & 56.53$_{\pm 1.80}$ \\
  De2  & 93.82$_{\pm 12.39}$ & 77.43$_{\pm 30.32}$ & 57.90$_{\pm 1.67}$ \\
  \bottomrule
  \end{tabular}
  \caption{Results of the robustness experiments in the \textsc{clevr} scenario. The results on three additional featured scenarios are included in Appendix~E.}
  \label{tab:robustness}
\end{table}

\paragraph{Heteromorphic populations.}
The next experiment assesses the applicability of the methodology to heteromorphic populations. Previous work has explored heterogeneity through differences in agents' internal design, such as architectural or learning asymmetries \citep{rita2022role, mahaut2025referential}. Here, we focus on perceptual heterogeneity as in \citet{degreeff2011development} and \citet{kim2021emergent}, testing whether a convention can still emerge when agents differ in what they can sense. For this purpose, we set up variations on the first four featured scenarios in which each individual agent has access to a randomly selected subset of the $d$ dimensions. Concretely, for every dataset, we run two instances of the experiment in which the agents are respectively endowed with combinations of $d-1$ and $\frac{d}{2}$ randomly selected sensors (He1 and He2 conditions). In order to establish a meaningful basis for comparison, we also run a version of the experiment with homomorphic populations in which the agents are endowed with the same number of sensors (Ho1 and Ho2).

As seen in Table~\ref{tab:robustness}, when moving from the homomorphic to the heteromorphic setting, we observe a similar trend as the one observed with the perceptual deviation experiments. While the conventionality of the language decreases, as agents will tend to use words that optimally fit their own sensory apparatus, they are still able to achieve a high degree of communicative success, even when equipped with a substantially different set of sensors.

\paragraph{Robustness against sensor defects.}
The last experiment validates the robustness of the methodology against sensor defects that occur in individual agents. We run a version of the experiment in which the agents suffer from a sudden malfunction after 500,000 interactions. Concretely, all agents suddenly lose access to one or half of their sensors (De1 and De2 conditions). The exact sensors that break down are randomly selected for each individual agent in each experimental run.

As seen in Table~\ref{tab:robustness}, the results show that the emergent languages become less conventional as the agents lose access to more sensors, but they still achieve high levels of communicative success. Using the He1 and He2 experiments as a basis for comparison, the experiment also demonstrates that the emergence of an effective linguistic convention prior to the malfunction remains beneficial even in the long term.

\subsection{Comparison with baseline and prior work}

To contextualise the proposed method, we compare it against a diagnostic baseline and the closest related approach from the literature. The baseline isolates the role of adaptive, interaction-driven concept formation.

\paragraph{Clustering baseline.}
As a diagnostic baseline, we combine $k$-means clustering \citep{lloyd1982least} with a standard naming game \citep{steels1995self}. This baseline replaces adaptive, interaction-driven concept formation with a fixed discretisation of the perceptual space: each agent partitions the space into a set of categories learned from the training data, after which agents negotiate labels over these categories through a naming game. We evaluate several values of $k \in \{10, 50, 100, 250, 500\}$. Performance improves as $k$ increases, with gains becoming marginal beyond $k = 100$. We report results with $k = 500$ as it yields the best performance. Full results are provided in Appendix~F.

In homogeneous populations, this baseline provides a useful reference point, as clustering yields a nearly shared discretisation of the perceptual space and the task reduces to negotiating labels over shared categories. In this setting, the coordination problem is therefore relatively straightforward and achieves 97\% success on the unseen test set. In contrast, when agents differ in their perceptions (i.e. conditions Ca1, Ca2, No1, No2, He1, He2, De1, De2 from Section~\ref{sec:robustness}), the learned clusters no longer align across agents. As a result, the labels cease to be as effective and communicative success collapses on the held-out test sets, ranging from 0.13\% to 43\%.

\paragraph{Comparison with neural approaches.}
To situate our methodology with respect to recent approaches to emergent communication on large-scale and high-dimensional inputs, we compare it with the closest comparable setup from the literature. As discussed in Section~\ref{sec:related-work}, most emergent communication experiments adopt a different game formulation, where the speaker observes only the topic while the listener observes the full scene. In some works, this setup has been referred to as the ``\textit{discrimination game}'' \citep[see e.g.][]{dessi2021interpretable,chaabouni2021communicating,chaabouni2022emergent, zion2024semantics}. In contrast, we consider a situated setting in which both agents share the same context, as defined in Section~\ref{sec:problem}. Only a small number of recent works study reference under such conditions, and these are typically restricted to two-agent setups and small synthetic datasets \citep{ohmer2022emergence,glowka2024contextdependent,kobrock2024context}.

We therefore compare our methodology against \citet{gualdoni2024bridging}, which ground their experiments in large-scale real-world image datasets and thus provide the closest available point of comparison for our approach. Their experiments use the ManyNames dataset \cite{silberer2020humans} which contains 25k real-world images annotated with free naming responses from native English speakers, spanning 442 distinct object names. Crucially, the annotations are never used for training, but serve only as a point of reference for analysing the emerged linguistic conventions. \citet{gualdoni2024bridging} construct scenes by pairing two entities detected within the same image, yielding pairs that are visually and semantically related. We follow the original processing pipeline of their paper.

\citet{gualdoni2024bridging} adopts the VQ-VIB framework of \citet{tucker2022trading} which formulate the emergence of lexical categories as a multi-objective optimisation problem that trades off utility, informativeness and complexity. In their setting, a purely utility-driven objective leads to large linguistic inventories of $\approx 959$ words, with smaller inventories emerging only when the three objectives are carefully balanced. In contrast, our approach is purely utility-driven and fully decentralised. In particular, we do not introduce any term that directly compares agents' internal representations, which is how informativeness is enforced in their framework. Despite this, when evaluated in the same setting, our method achieves higher communicative success on the held-out test set ($98.77\%$ vs.\ $95\%$) while producing substantially smaller inventories ($\approx 408$ words). These results suggest that, in a fully situated reference game with shared context, compact and effective conventions can emerge from interaction-level reward alone, without explicit regularisation terms. Finally, we evaluate the VQ-VIB approach in our experimental settings. As this approach is limited to two agents, we report results on populations of two agents and experiment with contexts of two  and ten entities (see detailed results in Appendix~F). In the \textit{two agent, two entity} setting, both approaches perform well, achieving 99.97\% (ours) and 95.29\% (VQ-VIB). However, when the difficulty of the task is increased to contexts with ten entities, the performance of VQ-VIB substantially degrades ($\le 42\%$), whereas our method maintains at least 82\% across all conditions.

\subsection{Multi-word utterances}

The results on homogeneous and heteromorphic populations in Sections~\ref{sec:homogeneous} and \ref{sec:robustness} demonstrate that the problem can be solved with single-word utterances ($\ell = 1$). When we allow longer utterances by setting $\ell > 1$, a trade-off emerges. Convergence becomes substantially slower: reaching the same communicative success achieved after 1M interactions with $\ell = 1$ requires more than 10M interactions when $\ell = 5$. At the same time, the resulting conventions are more conventional and substantially more compact. Across the seven featured scenarios, for $\ell = 1$, Table~\ref{tab:tabular-results} shows inventories ranging from 17 to 265 words. With $\ell = 5$, by contrast, the inventories fall to between 17 and 48 words (see Appendix~G). Agents converge on substantially smaller inventories because words can be reused across utterances rather than requiring many specialised single-word forms. Despite allowing multi-word utterances, agents still rely overwhelmingly on single-word utterances. During the early stages of training, roughly 80\% of interactions involve a single word. After 1M training interactions, the evaluation on the held-out test set shows that 95.39\% of utterances consist of a single word, 4.60\% contain two words, and fewer than 0.01\% contain more than two words.

The strong performance observed with $\ell = 1$ shows that, in the reference task, successful generalisation can be achieved without compositional (multi-word) utterances. This finding is consistent with previous work in the literature where, for example, \citet{chaabouni2020compositionality} demonstrate that in neural agents there is no correlation between the degree of compositionality and the ability to generalise. \citet{conklin2023compositionality} suggest that this lack of correlation arises because, once a language is sufficiently structured to support generalisation, additional regularity does not necessarily improve performance. In our approach, however, the ability to generalise without multi-word utterances follows from the open-ended nature of the inventories. To resolve communicative impasses, agents reshape existing words or invent new ones, progressively carving up the conceptual space. When restricted to single-word utterances, this pressure to distinguish meanings cannot be realised through combinations of smaller units and must instead be encoded in individual words. As a result, some words come to express combinations of features that would otherwise be distributed across multiple words. Over time, these words stabilise and occupy distinct niches, allowing populations to establish successful conventions without compositional utterances, as different words progressively fill the remaining gaps.

% --------------
% + CONCLUSION +
% --------------
\section{Conclusion}

This paper has introduced a fully decentralised methodology through which populations of autonomous agents converge on grounded linguistic conventions via local communicative interactions. An extensive evaluation on a diverse selection of 37 datasets, ranging from physico-chemical analyses to real-world images, confirmed its generality across environments. The proposed approach distinguishes itself from prior work by showing that effective global conventions can emerge from purely local interactions in population settings, without relying on shared representations, centralised training signals, perceptual discretisations or optimisation objectives beyond communicative success. It thereby brings together conditions that have typically been studied in isolation, in particular decentralisation, population-level dynamics, continuous perceptual grounding and perceptual heterogeneity. Not only are the resulting conventions highly effective at solving the reference task, they are also robust against perceptual variation, morphological differences between agents, and changing environmental conditions. At the same time, both the learning dynamics and the emergent languages remain transparent and human-interpretable, allowing the progressive structuring and alignment of conceptual spaces to be directly observed. Overall, our findings support the view that key properties of human language, most notably conventionality, robustness and adaptivity, can arise from usage-based mechanisms at the level of the local interaction alone, without the need for shared representations, central control or explicit regularisation objectives.

\section*{Acknowledgements}

We are grateful to Fabio De Ponte, Liesbet De Vos, Jonas Gillain, Tom Lenaerts, Arno Temmerman, Maxime Toquebiau, Remi van Trijp and Jamie Wright for their insightful comments on an earlier version of this manuscript, as well as to the three anonymous TACL reviewers and the TACL action editor Daniel Fried for their invaluable work and constructive suggestions that have led to a better paper. This research was supported by the F.R.S.-FNRS-FWO WEAVE project HERMES I under grant numbers T002724F (F.R.S.-FNRS) and G0AGU24N (FWO), the Flemish Government under the Onderzoeksprogramma Artificiële Intelligentie (AI) Vlaanderen programme and the AI Flagship project ARIAC by DigitalWallonia4.ai.

\bibliography{bibliography}
\bibliographystyle{acl_natbib}

\section*{Appendix}

\appendix 

\section{Hardware, training regime, tuned hyperparameters}
\label{appendix:training}
All experiments were conducted on 20-core INTEL Xeon Gold 6148 processors, paired with 32GB of RAM. One million sequential interactions (the amount of communicative interactions in each experiment) were executed on this hardware in $\pm$ 6 to 8 hours. Table \ref{tab:tuned-params-appendix} includes the space of hyperparameters explored for the baseline \textsc{clevr} experiment. The best performing set of hyperparameters (in terms of communicative success and linguistic conventionalisation) are reported in the main text (see Table 1 in the main text). Every subsequent experiment uses this same set of parameters.

\begin{table}[h!]
\small
\centering
\begin{tabular}{lcc}
\toprule
\textbf{Param.} & \textbf{Tested values} \\ 
\midrule
$s_r$ & $\{+0.01, +0.05, +0.1\}$ \\ 
$s_p$ & $\{-0.01, -0.05, -0.1\}$ \\ 
$s_{li}$ & $\{-0.05, -0.01, -0.02, -0.05, -0.1\}$ \\
\midrule
$\sigma_i$ & $\{0.001, 0.005, 0.01, 0.05, 0.1\}$ \\ 
$\omega_i$ & $\{0.1, 0.2, 0.5, 0.75, 1.0\}$ \\ 
$c_r$ & $\{+1, +5, +10\}$ \\ 
$c_p$ & $\{-1, -5, -10\}$ \\ 
\bottomrule 
\end{tabular} 
\caption{Overview of hyperparameter search}
\label{tab:tuned-params-appendix}
\end{table}

\section{Data processing pipelines}
\label{appendix:preprocesing-data}

\paragraph{Tabular datasets}

This paper uses 33 publicly available tabular datasets to validate the methodology. Each tabular dataset stores information in rows and columns, where rows represent entities and columns represent continuous or categorical features. The datasets can be broadly classified into one of three categories: (i) 7 contain only continuous features, (ii) 24 mix continuous and categorical features, (iii) and 2 contain only categorical features. The processing pipeline begins by removing columns containing all missing values and rows with any missing values. Duplicate rows are removed, keeping only the first occurrence. As some datasets represent discrete categorical information as integers, these `continuous' features are converted to categorical features. Next, all continuous features are normalised. Finally, the datasets are divided into training and test sets using a 75\%/25\% split.

\paragraph{CLEVR}

As described in the main text, the \textsc{clevr} scenario uses images from the CLEVR dataset \citep{johnson2017clevr}, processed following the method outlined by \citet{nevens2020continuous}. The dataset contains 85K images, each depicting 3 to 10 geometric objects. We retain the original data splits, with 70K images for training and 15K for testing. After processing, each depicted object is represented through a feature vector. These features correspond to information obtained through computer vision techniques (e.g. width-height ratio, colour channel values, x-axis position, etc.). All 20 dimensions of these feature vectors are continuously-valued and normalised.

\paragraph{Real-world images}
We use four computer vision datasets: \textsc{mscoco} (v. 2017) \citep{lin2014microsoft}, \textsc{celeba} \citep{liu2015deep}, Caltech-UCSD \textsc{birds} \citep{wah2011caltech} and \textsc{ManyNames} \cite{silberer2020humans}. \textsc{mscoco} (2017) contains over 159k naturalistic images of everyday scenes, \textsc{celeba} contains over 202k images of celebrity faces, Caltech-UCSD \textsc{birds} contains approximately 12k images covering 200 bird species and the \textsc{manynames} dataset consists of 25k real-world images of everyday scenes. For \textsc{mscoco} we use the original dataset splits, while for \textsc{celeba} and \textsc{birds} we use a 75\%/25\% train-test split. In contrast to the \textsc{clevr} setting, where agents communicate about objects within images, here each image is treated as a single entity and agents play language games about images as wholes. Each image is processed by a pre-trained DinoV2 model \citep{oquab2023dinov2}, yielding a 384-dimensional embedding that serves as the agent's perceptual input. The \textsc{manynames} dataset is used for comparison with \citet{gualdoni2024bridging}. We therefore follow, in this case, the exact processing pipeline of their paper.

\section{Results on the other 30 datasets}
\label{sec:results-appendix}

In Section~6.1 of the main text, the results on seven featured scenarios are presented. In Table \ref{tab:tabular-results-appendix}, we present the full evaluation on the other 30 scenarios.

\section{Effect of population size on convergence}
\label{sec:scaling-appendix}

In Table \ref{tab:scaling-appendix}, we test how the approach scales with population size by varying the number of $k$ agents while keeping all other settings fixed (see Table~1 in the main text) in the \textsc{clevr} scenario. We test populations of up to 100 agents ($k \in \{2,10,25,50,100\}$). The population is trained for 1M (but also up to 10M) interactions and evaluated on 100K interactions.

\section{Experimental results demonstrating robustness of methodology}

In Section~6.2 of the main text, four experiments are evaluated on \textsc{clevr} to demonstrate the robustness of our methodology (see paragraphs \textit{Uncalibrated sensors and noisy environments}, \textit{Heteromorphic populations}, \textit{Robustness against sensor defects}). In Table~\ref{tab:robustness-appendix}, we present the results on the other three featured scenarios: \textsc{winery}, \textsc{exoplanets}, and \textsc{mushrooms}.

\begin{table*}[h!]
\small
\centering
\begin{tabular}{p{2.5cm}p{0.9cm}p{1cm}p{0.9cm}p{1.7cm}p{1.7cm}p{1.85cm}p{1.7cm}}
\toprule
\textbf{Dataset} & \textbf{\# ent.} & \textbf{\# cont.} & \textbf{\# cat.} & \textbf{succ. (\%) }& \textbf{conv. (\%)} & \textbf{inv. size} & \textbf{inv. size (95\%)} \\
\midrule
\citet{jadikar2019gas}                      & 37K & 11 & 0      & 99.66$_{\pm 0.12}$ & 89.59$_{\pm 1.42}$ & 76.57$_{\pm 1.29}$ & 56.18$_{\pm 1.87}$ \\
\citet{devito2008field}                     & 7K & 15 & 0       & 99.66$_{\pm 0.15}$ & 91.00$_{\pm 1.07}$ & 77.77$_{\pm 1.20}$ & 58.06$_{\pm 2.04}$ \\
\citet{ma2019league}                        & 10K & 39 & 0      & 99.53$_{\pm 0.12}$ & 92.21$_{\pm 0.69}$ & 86.67$_{\pm 2.14}$ & 51.40$_{\pm 1.67}$ \\
\citet{vijaya2018steel}                     & 303 & 16 & 0      & 99.50$_{\pm 0.27}$ & 87.74$_{\pm 2.66}$ & 88.63$_{\pm 1.40}$ & 56.50$_{\pm 1.50}$ \\
\citet{tuameh2023physical}                  & 1K & 99 & 0       & 99.46$_{\pm 0.10}$ & 93.46$_{\pm 0.96}$ & 81.22$_{\pm 2.48}$ & 41.69$_{\pm 1.07}$ \\
\citet{brooks1989airfoil}                   & 2K & 6 & 0        & 99.14$_{\pm 0.21}$ & 90.11$_{\pm 2.01}$ & 116.89$_{\pm 3.19}$ & 59.38$_{\pm 1.43}$ \\
\midrule
\citet{dalpozzolo2014learned}               & 284K & 30 & 1     & 99.77$_{\pm 0.07}$ & 88.35$_{\pm 1.13}$ & 73.24$_{\pm 1.17}$ & 59.02$_{\pm 1.47}$ \\
\citet{boksha2024banana}                    & 8K & 7 & 1        & 99.71$_{\pm 0.14}$ & 90.35$_{\pm 1.14}$ & 65.49$_{\pm 1.20}$ & 49.26$_{\pm 1.16}$ \\
\citet{kadiwal2021water}                    & 2K & 9 & 1        & 99.69$_{\pm 0.15}$ & 89.51$_{\pm 1.36}$ & 66.60$_{\pm 1.09}$ & 49.99$_{\pm 1.70}$ \\
\citet{smith1988using}                      & 768 & 8 & 1       & 99.65$_{\pm 0.11}$ & 91.10$_{\pm 1.01}$ & 82.90$_{\pm 1.49}$ & 55.93$_{\pm 1.35}$ \\
\citet{olteanu2020gtzan}                    & 10K & 58 & 1      & 99.64$_{\pm 0.12}$ & 88.83$_{\pm 1.67}$ & 74.89$_{\pm 1.56}$ & 55.93$_{\pm 1.76}$ \\
\citet{koklu2020multiclass}                 & 14K & 16 & 1      & 99.57$_{\pm 0.25}$ & 92.58$_{\pm 1.51}$ & 78.54$_{\pm 1.63}$ & 47.89$_{\pm 1.34}$ \\
\citet{agrawal2017diamonds}                 & 54K & 7 & 3       & 99.56$_{\pm 0.11}$ & 93.47$_{\pm 0.88}$ & 95.50$_{\pm 2.12}$ & 57.70$_{\pm 1.83}$ \\
\citet{wolberg1993breast}                   & 569 & 31 & 1      & 99.53$_{\pm 0.19}$ & 91.34$_{\pm 1.27}$ & 75.42$_{\pm 2.66}$ & 47.96$_{\pm 1.48}$ \\
\citet{usda2023nutrients}                   & 5K & 66 & 1       & 99.52$_{\pm 0.21}$ & 89.19$_{\pm 1.59}$ & 82.23$_{\pm 2.18}$ & 56.41$_{\pm 1.56}$ \\
\citet{kottarathil2022bitcoin}              & 611K & 7 & 1      & 99.50$_{\pm 0.19}$ & 92.65$_{\pm 0.65}$ & 91.90$_{\pm 1.03}$ & 52.48$_{\pm 1.72}$ \\
\citet{sejnowski1988connectionist}          & 208 & 60 & 1      & 99.44$_{\pm 0.48}$ & 87.03$_{\pm 2.45}$ & 65.51$_{\pm 3.19}$ & 41.93$_{\pm 1.63}$ \\
\citet{lo2024fish}                          & 4K & 3 & 1        & 99.34$_{\pm 0.10}$ & 91.35$_{\pm 1.63}$ & 124.89$_{\pm 2.91}$ & 66.00$_{\pm 1.60}$ \\
\citet{er2024indian}                        & 51K & 126 & 2     & 99.34$_{\pm 0.59}$ & 89.14$_{\pm 2.53}$ & 82.30$_{\pm 1.45}$ & 56.22$_{\pm 1.37}$ \\
\citet{jikadara2024brand}                   & 1K & 10 & 10      & 98.94$_{\pm 0.23}$ & 89.64$_{\pm 2.20}$ & 146.78$_{\pm 2.02}$ & 64.18$_{\pm 1.78}$ \\
\citet{kaggle2024student}                   & 6K & 7 & 13       & 98.87$_{\pm 0.22}$ & 88.98$_{\pm 1.85}$ & 140.76$_{\pm 3.05}$ & 67.86$_{\pm 1.49}$ \\
\citet{romerohernandez2022customer}         & 2K & 18 & 10      & 98.83$_{\pm 0.32}$ & 92.34$_{\pm 1.25}$ & 89.67$_{\pm 1.59}$ & 43.68$_{\pm 1.49}$ \\
\citet{mujtaba2024color}                    & 765 & 3 & 1       & 98.78$_{\pm 0.34}$ & 91.68$_{\pm 1.52}$ & 87.15$_{\pm 1.52}$ & 41.95$_{\pm 1.82}$ \\
\citet{bart2015corgis}                      & 3K & 9 & 8        & 98.62$_{\pm 0.62}$ & 92.58$_{\pm 2.79}$ & 107.90$_{\pm 2.23}$ & 46.10$_{\pm 1.79}$ \\
\citet{francois2024worlds}                  & 1K & 4 & 3        & 98.50$_{\pm 0.41}$ & 93.65$_{\pm 2.29}$ & 118.48$_{\pm 2.87}$ & 48.56$_{\pm 1.32}$ \\
\citet{khorasani2024electric}               & 1K & 10 & 7       & 98.40$_{\pm 0.31}$ & 93.87$_{\pm 0.57}$ & 84.78$_{\pm 2.69}$ & 33.18$_{\pm 1.43}$ \\
\citet{ms2024pokemon}                       & 1K & 7 & 2        & 97.42$_{\pm 1.43}$ & 90.67$_{\pm 1.63}$ & 93.27$_{\pm 1.03}$ & 42.75$_{\pm 2.32}$ \\
\citet{fisher1936iris}                      & 147 & 4 & 1       & 97.27$_{\pm 0.59}$ & 80.52$_{\pm 2.12}$ & 91.68$_{\pm 2.56}$ & 48.61$_{\pm 0.95}$ \\
\citet{banik2018pokemon}                    & 339 & 33 & 5      & 95.60$_{\pm 1.69}$ & 86.96$_{\pm 3.02}$ & 84.62$_{\pm 2.18}$ & 40.70$_{\pm 2.44}$ \\
\midrule
\citet{bohanec1998knowledge}                & 2K & 0 & 7        & 99.08$_{\pm 0.15}$ & 92.02$_{\pm 1.04}$ & 139.61$_{\pm 3.89}$ & 42.08$_{\pm 1.27}$ \\
\bottomrule
\end{tabular}
\caption{Experimental results on the held-out test sets of the 30 remaining scenarios. Mean and 2 standard deviations computed over 10 independent experimental runs. The columns describe the dataset, number of entities, number of continuous dimensions, number of categorical dimensions, communicative success, conventionality and linguistic inventory size.}
\label{tab:tabular-results-appendix}
\end{table*}

\begin{table*}
\small
\centering
\begin{tabular}{llcccc}
\toprule
\textbf{Dataset} & \textbf{Condition} & \textbf{succ. (\%)} & \textbf{conv. (\%)} & \textbf{inv. size} & \textbf{inv. size (95\%)}\\
\midrule
\multicolumn{5}{l}{\textbf{Uncalibrated sensors and noisy environments}} \\
\midrule
\textsc{wine}       & Baseline  & 99.76$_{\pm 0.17}$ & 88.34$_{\pm 1.31}$ & 77.12$_{\pm 1.21}$ & 60.25$_{\pm 1.37}$ \\
\textsc{wine}       & Ca1       & 99.71$_{\pm 0.15}$ & 88.83$_{\pm 1.89}$ & 77.95$_{\pm 1.34}$ & 58.96$_{\pm 1.34}$ \\
\textsc{wine}       & Ca2       & 99.58$_{\pm 0.21}$ & 88.04$_{\pm 1.66}$ & 77.94$_{\pm 2.14}$ & 57.75$_{\pm 2.14}$ \\
\textsc{wine}       & No1       & 97.61$_{\pm 0.46}$ & 72.73$_{\pm 2.51}$ & 68.89$_{\pm 0.82}$ & 53.92$_{\pm 1.69}$ \\
\textsc{wine}       & No2       & 76.89$_{\pm 2.46}$ & 38.17$_{\pm 2.22}$ & 78.74$_{\pm 3.64}$ & 37.47$_{\pm 2.13}$ \\
\midrule
\textsc{exoplanets} & Baseline  & 99.67$_{\pm 0.10}$ & 92.30$_{\pm 0.86}$ & 79.35$_{\pm 1.18}$ & 54.47$_{\pm 1.67}$ \\
\textsc{exoplanets} & Ca1       & 99.46$_{\pm 0.29}$ & 90.98$_{\pm 1.27}$ & 79.79$_{\pm 1.77}$ & 54.71$_{\pm 1.33}$ \\
\textsc{exoplanets} & Ca2       & 97.93$_{\pm 1.22}$ & 87.95$_{\pm 3.06}$ & 85.98$_{\pm 1.21}$ & 51.45$_{\pm 1.42}$ \\
\textsc{exoplanets} & No1       & 94.23$_{\pm 0.88}$ & 69.54$_{\pm 2.38}$ & 74.25$_{\pm 1.66}$ & 46.38$_{\pm 1.49}$ \\
\textsc{exoplanets} & No2       & 68.89$_{\pm 1.86}$ & 44.54$_{\pm 1.94}$ & 134.95$_{\pm 6.81}$ & 30.25$_{\pm 1.00}$ \\
\midrule
\multicolumn{5}{l}{\textbf{Heteromorphic populations}} \\
\midrule
\textsc{wine}       & Ho1       & 99.73$_{\pm 0.08}$  & 88.71$_{\pm 1.01}$ & 77.62$_{\pm 1.39}$ & 58.79$_{\pm 1.41}$    \\
\textsc{wine}       & He1       & 93.47$_{\pm 2.35}$  & 77.46$_{\pm 4.22}$ & 77.65$_{\pm 1.39}$ & 56.58$_{\pm 1.86}$    \\
\textsc{wine}       & Ho2       & 99.62$_{\pm 0.23}$  & 89.81$_{\pm 2.04}$ & 86.18$_{\pm 4.92}$ & 56.28$_{\pm 2.63}$    \\
\textsc{wine}       & He2       & 63.51$_{\pm 4.36}$  & 41.10$_{\pm 4.97}$ & 110.52$_{\pm 6.53}$ & 36.38$_{\pm 1.34}$   \\
\midrule
\textsc{exoplanets} & Ho1       & 99.59$_{\pm 0.33}$  & 92.89$_{\pm 1.43}$ & 82.46$_{\pm 5.30}$ & 50.90$_{\pm 1.35}$    \\
\textsc{exoplanets} & He1       & 93.23$_{\pm 4.83}$  & 78.32$_{\pm 9.00}$ & 84.16$_{\pm 2.46}$ & 50.43$_{\pm 1.63}$    \\
\textsc{exoplanets} & Ho2       & 96.31$_{\pm 6.10}$  & 91.55$_{\pm 12.46}$ & 151.20$_{\pm 10.37}$ & 40.87$_{\pm 5.83}$ \\
\textsc{exoplanets} & He2       & 18.29$_{\pm 30.45}$ & 8.15$_{\pm 23.67}$ & 211.97$_{\pm 30.45}$ & 84.04$_{\pm 27.02}$ \\
\midrule
\textsc{mushrooms}  & Ho1       & 97.78$_{\pm 1.17}$  & 86.07$_{\pm 2.90}$ & 270.08$_{\pm 16.32}$ & 74.92$_{\pm 3.07}$  \\
\textsc{mushrooms}  & He1       & 95.04$_{\pm 1.91}$  & 80.66$_{\pm 3.17}$ & 285.29$_{\pm 9.13}$ & 76.57$_{\pm 2.74}$   \\
\textsc{mushrooms}  & Ho2       & 88.97$_{\pm 7.13}$  & 81.70$_{\pm 5.56}$ & 180.72$_{\pm 34.70}$ & 60.26$_{\pm 3.68}$  \\
\textsc{mushrooms}  & He2       & 53.37$_{\pm 4.93}$  & 30.56$_{\pm 5.59}$ & 176.84$_{\pm 5.53}$ & 73.16$_{\pm 2.33}$   \\
\midrule
\multicolumn{5}{l}{\textbf{Robustness against sensor defects}} \\
\midrule
\textsc{wine}   	& De1       & 97.72$_{\pm 3.86}$    & 84.18$_{\pm 5.48}$    & 78.44$_{\pm 1.39}$ & 56.12$_{\pm 2.52}$   \\
\textsc{wine}   	& De2 	    & 88.46$_{\pm 23.29}$   & 69.27$_{\pm 36.39}$   & 84.17$_{\pm 5.82}$ & 53.70$_{\pm 1.35}$   \\
\midrule
\textsc{exoplanets}	& De1       & 97.69$_{\pm 4.67}$    & 87.18$_{\pm 8.53}$    & 80.80$_{\pm 1.94}$ & 51.12$_{\pm 2.38}$   \\
\textsc{exoplanets}	& De2	    & 73.37$_{\pm 55.87}$   & 60.92$_{\pm 62.47}$   & 102.84$_{\pm 23.96}$ & 51.22$_{\pm 2.53}$ \\
\midrule
\textsc{mushrooms}	& De1       & 96.75$_{\pm 1.70}$    & 83.95$_{\pm 5.73}$    & 257.68$_{\pm 12.71}$ & 75.38$_{\pm 3.02}$ \\
\textsc{mushrooms} 	& De2	    & 83.53$_{\pm 28.77}$   & 66.04$_{\pm 41.85}$   & 225.97$_{\pm 25.49}$ & 72.96$_{\pm 1.96}$ \\
\bottomrule
\end{tabular}
\caption{Results of the robustness experiments on the held-out test set in Section~5.3 of the main paper on the \textsc{clevr}, \textsc{wine} and \textsc{exoplanets} and \textsc{mushrooms} datasets. Mean and 2 standard deviations computed over 10 independent runs.}
\label{tab:robustness-appendix}
\end{table*}

\begin{table*}
\small
\centering
\begin{tabular}{lcccc}
\toprule
\textbf{\# agents} & \textbf{succ. (\%)} & \textbf{conv. (\%)} & \textbf{inv. size} & \textbf{inv. size (95\%)} \\
\midrule
\multicolumn{5}{l}{\textbf{1M interactions}} \\
\midrule
$k = 2$ & 100.00$_{\pm 0.00}$ & 100.00$_{\pm 0.00}$ & 60.40$_{\pm 2.14}$ & 40.40$_{\pm 1.65}$ \\
$k = 10$ & 99.76$_{\pm 0.08}$ & 94.41$_{\pm 1.46}$ & 52.63$_{\pm 1.73}$ & 31.14$_{\pm 1.11}$ \\
$k = 25$ & 99.01$_{\pm 0.45}$ & 91.75$_{\pm 1.95}$ & 49.09$_{\pm 1.21}$ & 25.33$_{\pm 0.48}$ \\
$k = 50$ & 97.79$_{\pm 1.19}$ & 89.28$_{\pm 3.52}$ & 45.71$_{\pm 2.29}$ & 21.34$_{\pm 1.26}$ \\
$k = 100$ & 92.69$_{\pm 1.96}$ & 81.14$_{\pm 2.94}$ & 45.61$_{\pm 1.74}$ & 17.99$_{\pm 0.71}$ \\
\midrule
\multicolumn{5}{l}{\textbf{10M interactions}} \\
\midrule
$k = 10$ & 99.97$_{\pm 0.02}$ & 97.28$_{\pm 0.76}$ & 68.36$_{\pm 1.79}$ & 40.49$_{\pm 1.65}$ \\
$k = 100$ & 99.16$_{\pm 0.42}$ & 93.39$_{\pm 1.45}$ & 68.66$_{\pm 5.10}$ & 26.23$_{\pm 1.02}$ \\
\bottomrule
\end{tabular}
\caption{Effect of population size in the \textsc{clevr} scenario. All runs are trained for 1M interactions except the two final rows, which are trained for 10M interactions. Mean and 2 standard deviations computed over 10 independent runs.}
\label{tab:scaling-appendix}
\end{table*}

\section{Multi-word utterances}

In Table \ref{tab:multi-word-appendix}, we test how the results vary when we vary the utterance length to maximally 5 words (i.e. $\ell = 5$). The population is trained for 10M interactions and evaluated for 100K interactions.

\begin{table*}
\small
\centering
\begin{tabular}{lrrrrrrrr}
\toprule
\textbf{Dataset} & \textbf{\# ent.} & \textbf{\# cont.} & \textbf{\# cat.} & \textbf{succ. (\%) }& \textbf{conv. (\%)} & \textbf{inv. size} & \textbf{inv. size (95\%)} \\
\midrule
\textsc{clevr}         & 468K  & 20    & 0   & 99.66$_{\pm 0.16}$ & 98.73$_{\pm 0.13}$ & 17.10$_{\pm 1.15}$ & 15.50$_{\pm 0.56}$ \\
\textsc{wine}          & 5K    & 12    & 0   & 99.59$_{\pm 0.35}$ & 98.05$_{\pm 1.29}$ & 30.67$_{\pm 1.76}$ & 26.00$_{\pm 1.00}$ \\
\textsc{exoplanets}    & 5K    & 8     & 4   & 99.14$_{\pm 0.56}$ & 96.53$_{\pm 3.21}$ & 31.53$_{\pm 1.12}$ & 25.57$_{\pm 1.69}$ \\
\textsc{mushrooms}     & 8K    & 0     & 23  & 99.05$_{\pm 0.98}$ & 92.72$_{\pm 0.96}$ & 48.17$_{\pm 3.70}$ & 31.23$_{\pm 1.40}$ \\
\midrule
\textsc{mscoco}        & 159K  & 384   & 0   & 98.03$_{\pm 0.62}$ & 98.56$_{\pm 0.60}$ & 21.83$_{\pm 1.58}$ & 17.75$_{\pm 0.72}$ \\
\textsc{celeba}        & 203K  & 384   & 0   & 98.41$_{\pm 0.67}$ & 98.12$_{\pm 2.05}$ & 26.01$_{\pm 2.72}$ & 19.24$_{\pm 1.36}$ \\
\textsc{birds}         & 12K   & 384   & 0   & 97.97$_{\pm 0.68}$ & 98.43$_{\pm 0.56}$ & 21.14$_{\pm 0.89}$ & 17.48$_{\pm 0.69}$ \\
\bottomrule
\end{tabular}
\caption{Experimental results on the held-out test set of the seven featured scenarios after training agents for 10M interactions, when agents are allowed to produce multi-word utterances ($\ell=5$). Mean and 2 standard deviations computed over 10 independent runs. The columns describe the dataset, number of entities, number of continuous and categorical dimensions, communicative success, conventionality and linguistic inventory size.}
\label{tab:multi-word-appendix}
\end{table*}

\section{Baseline and comparison with prior work}

In Table \ref{tab:baselines-appendix}, we evaluate a clustering-based baseline and a recent neural emergent communication approach (VQ-VIB) \cite{gualdoni2024bridging} in the \textsc{wine} scenario. For our approach and the clustering baseline population is trained for 1M interactions and evaluated one 100K interactions. For VQ-VIB, we run the released code of \citet{gualdoni2024bridging} on the \textsc{wine} dataset using their best-performing configuration, which optimises solely for utility. We evaluate VQ-VIB with population of two agents, as their framework only supports this setting, and consider scenes containing two and ten entities.

\begin{table*}
  \small
  \centering
  \setlength{\tabcolsep}{2.5pt}
  \begin{tabular}{c r r r r r r r r }
  \toprule
  & \multicolumn{3}{c}{\textbf{This work}}
  & \multicolumn{3}{c}{\textbf{Clustering}}
  & \multicolumn{2}{c}{\textbf{VQ-VIB}} \\
  \cmidrule(lr){2-4} \cmidrule(lr){5-7} \cmidrule(lr){8-9}
  \textbf{Setting} 
  & \textbf{2-2} & \textbf{2-10} & \textbf{10-10}
  & \textbf{2-2} & \textbf{2-10} & \textbf{10-10}
  & \textbf{2-2} & \textbf{2-10} \\
  \midrule
  Baseline     & 99.97$_{\pm 0.02}$ & 100.00$_{\pm 0.00}$ & 99.52$_{\pm 0.19}$     & 99.74$_{\pm 0.05}$    & 97.60$_{\pm 0.16}$    & 97.60$_{\pm 0.18}$  & 95.29$_{\pm 1.86}$ & 47.33$_{\pm 23.41}$  \\
  Ca1         & 99.77$_{\pm 0.09}$ & 99.69$_{\pm 0.22}$ & 98.81$_{\pm 0.39}$      & 39.11$_{\pm 3.86}$    & 38.33$_{\pm 3.60}$    & 28.58$_{\pm 1.14}$  & 94.57$_{\pm 1.73}$ & 32.28$_{\pm 27.48}$  \\
  Ca2         & 96.69$_{\pm 1.99}$ & 89.46$_{\pm 4.80}$ & 80.02$_{\pm 4.01}$      & 37.52$_{\pm 2.36}$    & 36.99$_{\pm 3.49}$    & 27.57$_{\pm 1.27}$  & 93.94$_{\pm 2.14}$ & 29.81$_{\pm 24.34}$  \\
  No1         & 99.72$_{\pm 0.13}$ & 99.46$_{\pm 0.13}$ & 98.99$_{\pm 0.25}$      & 39.92$_{\pm 3.05}$    & 38.36$_{\pm 2.87}$    & 28.46$_{\pm 1.06}$  & 92.79$_{\pm 3.00}$ & 42.89$_{\pm 26.56}$  \\
  No2         & 96.09$_{\pm 1.08}$ & 82.25$_{\pm 1.33}$ & 83.62$_{\pm 1.06}$      & 37.06$_{\pm 3.56}$    & 36.72$_{\pm 2.96}$    & 27.42$_{\pm 0.98}$  & 87.21$_{\pm 6.03}$ & 38.88$_{\pm 26.12}$  \\
  Ho1         & 99.97$_{\pm 0.02}$ & 100.00$_{\pm 0.00}$ & 99.51$_{\pm 0.11}$     & 39.87$_{\pm 4.39}$    & 39.33$_{\pm 4.34}$    & 28.83$_{\pm 2.37}$  & 93.45$_{\pm 3.27}$ & 47.13$_{\pm 26.64}$  \\
  Ho2         & 99.99$_{\pm 0.01}$ & 99.99$_{\pm 0.01}$ & 99.29$_{\pm 0.30}$      & 42.36$_{\pm 6.27}$    & 41.68$_{\pm 5.23}$    & 32.04$_{\pm 6.11}$  & 93.84$_{\pm 1.27}$ & 45.69$_{\pm 20.71}$  \\

  He1         & 97.18$_{\pm 2.87}$ & 99.92$_{\pm 0.18}$ & 93.66$_{\pm 2.91}$      & 29.17$_{\pm 12.69}$   & 28.98$_{\pm 11.83}$   & 18.76$_{\pm 3.93}$  & 93.87$_{\pm 2.66}$ & 39.92$_{\pm 18.11}$  \\
  He2         & 84.35$_{\pm 12.75}$ & 98.27$_{\pm 2.71}$ & 64.10$_{\pm 4.42}$     & 8.52$_{\pm 6.22}$     & 7.92$_{\pm 5.09}$     & 3.97$_{\pm 1.07}$   & 91.05$_{\pm 2.38}$ & 34.42$_{\pm 26.83}$  \\
  De1         & 98.57$_{\pm 1.94}$ & 99.89$_{\pm 0.32}$ & 96.29$_{\pm 2.20}$      & 37.09$_{\pm 5.13}$    & 33.26$_{\pm 3.79}$    & 25.35$_{\pm 4.00}$  & 93.55$_{\pm 3.26}$ & 43.76$_{\pm 27.05}$  \\
  De2         & 88.64$_{\pm 4.29}$ & 99.43$_{\pm 0.44}$ & 80.79$_{\pm 3.36}$      & 14.80$_{\pm 7.94}$    & 13.03$_{\pm 6.29}$    & 12.54$_{\pm 1.60}$  & 86.38$_{\pm 5.24}$ & 38.54$_{\pm 29.31}$  \\
  \bottomrule
  \end{tabular}
  \caption{Comparison against the clustering baseline k-means \cite{lloyd1982least} and the neural VQ-VIB approach \cite{gualdoni2024bridging}. Results on held-out test sets in the \textsc{wine} scenario. Mean and 2 standard deviations computed over ten independent runs. In the column labels, $x$-$y$ denotes a setting with $x$ agents and scenes containing $y$ entities, where one entity is the topic and the others act as distractors.}
  \label{tab:baselines-appendix}
\end{table*}

\end{document}